\documentclass[authoryear]{elsarticle}

\usepackage{amsmath,amsfonts,amssymb}
\usepackage{graphicx,epsfig}
\usepackage{booktabs,xtab}
\usepackage[]{natbib}
\usepackage[ruled,lined,vlined]{algorithm2e}
\usepackage{pseudocode}
\usepackage{booktabs,xtab,longtable,multirow,multicol}
\usepackage[small,center,bf,singlelinecheck=off]{caption}
\usepackage{colortbl}
\usepackage{lscape,afterpage}

\def\nbnewclosed{$12$}
\def\nbsolved{$300$}

\newdefinition{defn}{Definition}
\newtheorem{prop}[defn]{Proposition}
\newproof{proof}{Proof}

\DeclareMathOperator{\LB}{LB} 
\DeclareMathOperator{\UB}{UB} 
\DeclareMathOperator{\PFT}{P} 
\DeclareMathOperator{\SOL}{SOL} 
\DeclareMathOperator{\Opt}{Opt} 
\DeclareMathOperator{\D}{D} 
\DeclareMathOperator{\TM}{TM} 
\DeclareMathOperator{\MIPS}{MIPS} 
\DeclareMathOperator{\MinLen}{MinLength} 

\newcommand{\VarOriginal}{\ensuremath{\mathrm{ORG}}\xspace} 
\newcommand{\VarStep}{\ensuremath{\mathrm{Step}}\xspace} 
\newcommand{\VarMandatory}{\ensuremath{M}\xspace} 

\usepackage{natbib}
 \bibpunct[, ]{(}{)}{,}{a}{}{,}%
 \def\bibsep{\smallskipamount}%
\begin{document}

\title{Solving the Team Orienteering Problem\\ with Cutting Planes}

\author[UTC,UL]{Racha El-Hajj \corref{cor1}} \ead{racha.el-hajj@hds.utc.fr}
\author[UN]{Duc-Cuong Dang} \ead{duc-cuong.dang@nottingham.ac.uk}
\author[UTC]{Aziz Moukrim} \ead{aziz.moukrim@hds.utc.fr}

\cortext[cor1]{corresponding author}
\address[UTC]{Sorbonne universit\'es, Universit\'e de technologie de Compi\`egne\\
CNRS, Heudiasyc UMR 7253, CS 60 319, 60 203 Compi\`egne cedex}
\address[UN]{University of Nottingham, School of Computer Science\\
Jubilee Campus, Wollaton Road, Nottingham NG8 1BB, United Kingdom}
\address[UL]{Universit\'{e} Libanaise, \'{E}cole Doctorale des Sciences et de Technologie\\
Campus Hadath, Beyrouth, Liban}

\begin{abstract}
The Team Orienteering Problem (TOP) is an attractive variant of the Vehicle Routing Problem (VRP). The aim is to select customers and at the same time organize the visits for a vehicle fleet so as to maximize the collected profits and subject to a travel time restriction on each vehicle. In this paper, we investigate the effective use of a linear formulation with polynomial number of variables to solve TOP. Cutting planes are the core components of our solving algorithm. It is first used to solve smaller and intermediate models of the original problem by considering fewer vehicles. Useful information are then retrieved to solve larger models, and eventually reaching the original problem. Relatively new and dedicated methods for TOP, such as identification of irrelevant arcs and mandatory customers, clique and independent-set cuts based on the incompatibilities, and profit/customer restriction on subsets of vehicles, are introduced. We evaluated our algorithm on the standard benchmark of TOP. The results show that the algorithm is competitive and is able to prove the optimality for \nbnewclosed{} instances previously unsolved.
\end{abstract}


\maketitle

\section{Introduction}\label{sec:description}

The Team Orienteering Problem (TOP) was first mentioned in \cite{bib:Butt94} as the Multiple Tour Maximum Collection Problem (MTMCP). Later, the term TOP was formally introduced in \cite{bib:Chao96a}. TOP is a variant of the Vehicle Routing Problem (VRP) \citep{bib:Archetti14}. In this variant, a limited number of identical vehicles is available to visit customers from a potential set. Two particular depots, the \emph{departure} and the \emph{arrival} points are considered. Each vehicle must perform its route starting from the departure depot and returning to the arrival depot without exceeding its predefined travel time limit. A certain amount of profit is associated for each customer and must be collected at most once by the fleet of vehicles. The aim of solving TOP is to organize an itinerary of visits respecting the above constraints for the fleet in such a way that the total amount of collected profits from the visited customers is maximized.

A special case of TOP is the one with a single vehicle. The resulted problem is known as the Orienteering Problem (OP), or the Selective Travelling Salesman Problem (STSP) (see the surveys by \citealp{bib:Feillet05}, \citealp{bib:Vansteenwegen11} and \citealp{bib:Gavalas14}). OP/STSP is already NP-Hard \citep{bib:Laporte90}, and so is TOP \citep{bib:Chao96a}. The applications of TOP arise in various situations. For example in \citet{bib:Bouly08a}, the authors used TOP to model the schedule of inspecting and repairing tasks in water distribution. Each task in this case has a specific level of urgency which is similar to a profit.
Due to the limitation of available human and material resources, the efficient selection of tasks as well as the route planning become crucial to the quality of the schedule. A very similar application was described in \citet{bib:Tang05} to route technicians to repair sites. In \citet{bib:Souffriau08}, \citet{bib:Vansteenwegen09b} and \citet{bib:Gavalas14}, the tourist guide service that offers to the customers the possibility to personalize their trips is discussed as variants of TOP/OP. In this case, the objective is to maximize the interest of customers on attractive places subject to their duration of stay. Those planning problems are called Tourist Trip Design Problems (TTDPs). Many other applications include the team-orienteering sport game, bearing the original name of TOP, the home fuel delivery problem with multiple vehicles \citep[e.g.,][]{bib:Chao96a} and the athlete recruiting from high schools for a college team \citep[e.g.,][]{bib:Butt94}.

Many heuristics have been proposed to solve TOP, like the ones in \citet{bib:Archetti07}, \citet{bib:Souffriau10}, \citet{bib:Dang13b} and \citet{bib:Kim13a}. These approaches are able to construct solutions of good quality in short computational times, but those solutions are not necessarily optimal. In order to validate them and evaluate the performance of the heuristic approaches, either optimal solutions or upper bounds are required. For this reason, some researches have been dedicated to elaborate exact solution methods for TOP. \citet{bib:Butt99} introduced a procedure based on the set covering formulation. A column generation algorithm was developed to solve this problem. In \citet{bib:Boussier07}, the authors proposed a branch-and-price (B-P) algorithm in which they used a dynamic programming approach to solve the pricing problem. Their approach has the advantage of being easily adaptable to different variants of the problem. Later, \citet{bib:Aragao10} introduced a pseudo-polynomial linear model for TOP and proposed a branch-cut-and-price (B-C-P) algorithm. New classes of inequalities, including min-cut and triangle clique, were added to the model and the resulting formulation was solved using a column generation approach. Afterwards, \cite{bib:Dang13a} proposed a branch-and-cut (B-C) algorithm based on a linear formulation and features a new set of valid inequalities and dominance properties in order to accelerate the solution process. Recently, \citet{bib:Keshtkarana15} proposed a Branch-and-Price algorithm with two relaxation stages (B-P-2R) and a Branch-and-Cut-and-Price (B-C-P) approach to solve TOP, where a bounded bidirectional dynamic programming algorithm with decremental state space relaxation was used to solve the subproblems. These five methods were able to prove the optimality for a large part of the standard benchmark of TOP \citep{bib:Chao96a}, however there is a large number of instances that are still open until now. Furthermore, according to the recent studies of \citet{bib:Dang13b} and \citet{bib:Kim13a}, it appears that it is hardly possible to improve the already-known solutions for the standard benchmark of TOP using heuristics. These studies suggest that the known heuristic solutions could be optimal but there is a lack of variety of effective methods to prove their optimality.

Motivated by the above facts, in this paper we propose a new exact algorithm to solve TOP. It is based on a linear formulation with a polynomial number of binary variables. Our algorithmic scheme is a cutting plane algorithm which exploits integer solutions of successive models with the \emph{subtour} elimination constraints being relaxed at first and then iteratively reinforced. Recently, \cite{bib:Pferschy13} demonstrates on the Travelling Salesman Problem (TSP) that such a technique which was almost forgotten could be made efficient nowaday with the impressive performance of modern solvers for Mixed-Integer Programming (MIP), especially with a careful control over the reinforcing of the subtour elimination. Our approach is similar but in addition to subtour elimination, we also make use of other valid inequalities and useful dominance properties to enhance the intermediate models. The properties include breaking the symmetry and exploiting bounds or optimal solutions of smaller instances/models with fewer number of vehicles, while the proposed valid inequalities are the clique cuts and the independent set cuts based on the incompatibilities between customers and between arcs. In addition, bounds on smaller restricted models are used to locate mandatory customers and inaccessible customers/arcs. Some of these cuts were introduced and tested in \cite{bib:Dang13a} yielding some interesting results for TOP, this encourages us to implement them immediately in our cutting plane algorithm. We evaluated our algorithm on the standard benchmark of TOP. The obtained results clearly show the competitiveness of our algorithm. The algorithm is able to prove the optimality for \nbnewclosed{} instances that none of the previous exact algorithms had been able to solve.

The remainder of the paper is organized as follows. A short description of the problem with its mathematical formulation is first given in Section \ref{sec:lp}, where the use of the generalized subtour elimination constraints is also discussed. In Section~\ref{sec:domprop}, the set of dominance properties, which includes symmetry breaking, removal of irrelevant components, identification of mandatory customers and boundaries on profits/numbers of customers, is presented. The graphs of incompatibilities between variables are also described in this section, along with the clique cuts and the independent set cuts. In Section \ref{sec:CuttingPlanegc}, all the techniques used to generate these efficient cuts are detailed, and the pseudocode of the main algorithmic scheme is given. Finally, the numerical results are discussed in Section \ref{sec:numresult}, and some conclusions are drawn.

\section{Problem formulation}\label{sec:lp}

TOP is modeled with a complete directed graph $G=(V, A)$ where $V =\{1,\dots,n\} \cup \{d, a\}$ is the set of vertices representing the customers and the depots, and $A=\{(i,j) \mid i,j \in V, i\neq j\}$ the set of arcs linking the different vertices together. The departure and the arrival depots for the vehicles are represented by the vertices $d$ and $a$. For convenience, we use the three sets $V^{-}$, $V^{d}$ and $V^{a}$ to denote respectively the sets of the customers only, of the customers with the departure depot and of the customers with the arrival one. A profit $p_i$ is associated for each vertex $i$ and is considered zero for the two depots ($p_d = p_a = 0$). Each arc $(i,j) \in A$ is associated with a travel cost $c_{ij}$. Theses costs are assumed to be symmetric and to satisfy the triangle inequality. All arcs incoming to the departure depot and outgoing from the arrival one must not be considered ($c_{id}=c_{ai}=\infty, \forall i \in V^{-}$). Let $F$ represent the fleet of the $m$ identical vehicles available to visit customers. Each vehicle must start its route from $d$, visit a certain number of customers and return to $a$ without exceeding its predefined travel cost limit $L$. Using these definitions, we can formulate TOP with a linear Mixed Integer Program (MIP) using a polynomial number of decision variables $y_{ir}$ and $x_{ijr}$. Variable $y_{ir}$ is set to $1$ if vehicle $r$ has served client $i$ and to $0$ otherwise, while variable $x_{ijr}$ takes the value $1$ when vehicle $r$ uses arc $(i,j)$ to serve customer $j$ immediately after customer $i$ and $0$ otherwise.

\begin{align}
\textrm{max} \sum_{i \in V^{-}}\sum_{r \in F} y_{ir} p_i \label{miptop:obj}\\
\sum_{r \in F} y_{ir} \leq 1 &\quad \forall i\in V^{-} \label{miptop:ctvisit}\\
\sum_{j \in V^a} x_{djr} = \sum_{j \in V^d} x_{jar} = 1 &\quad \forall r \in F \label{miptop:ctdepot}\\
\sum_{i \in V^a \setminus \{k\}} x_{kir} = \sum_{j \in V^d \setminus \{k\}} x_{jkr} = y_{kr} &\quad \forall k \in V^{-}, \forall r \in F \label{miptop:ctlink}\\
\sum_{i \in V^d}\sum_{j \in V^a \setminus \{i\}} c_{ij} x_{ijr} \leq L &\quad \forall r \in F \label{miptop:ctlength}\\
\sum_{(i,j)\in U \times U} x_{ijr} \leq |U| - 1 &\quad \forall U \subseteq V^{-}, |U| \geq 2, \forall r \in F \label{miptop:ctsubtour}\\
x_{ijr} \in \{0, 1\} &\quad \forall i \in V, \forall j \in V, \forall r \in F \label{miptop:ctinteger}\\
y_{ir} \in \{0, 1\} &\quad \forall i \in V^{-}, \forall r \in F \nonumber
\end{align}

The objective function \eqref{miptop:obj} maximizes the sum of collected profits from the visited customers. Constraints \eqref{miptop:ctvisit} impose that each customer must be visited at most once by one vehicle. Constraints \eqref{miptop:ctdepot} guarantee that each vehicle starts its path at vertex $d$ and ends it at vertex $a$, while constraints \eqref{miptop:ctlink} ensure the connectivity of each tour. Constraints \eqref{miptop:ctlength} are used to impose the travel length restriction, while constraints \eqref{miptop:ctsubtour} eliminate all possible subtours, i.e. cycles excluding the depots, from the solution. Finally, constraints \eqref{miptop:ctinteger} set the integral requirement on the variables.

Enumerating all constraints \eqref{miptop:ctsubtour} yields a formulation with an exponential number of constraints. In practice, these constraints are first relaxed from the formulation, then only added to the model whenever needed. The latter can be detected with the presence of subtours in the solution of the relaxed model. We also replace constraints \eqref{miptop:ctsubtour} with the stronger ones, the so-called Generalized Subtour Elimination Constraints (GSECs) which enhance both the elimination of specific subtours and the connectivity in the solution. The first GSEC experiment with OP were reported in \citet{bib:Fischetti98}.

We adapted the GSEC version from \citet{bib:Dang13a} formulated to TOP with a directed graph as follows. For a given subset $S$ of customer vertices, we define $\delta(S)$ to be the set of arcs that connect vertices in $S$ with those outside $S$, i.e. vertices in $V \setminus S$. We also use $\gamma(S)$ to represent the set of arcs interconnecting vertices in $S$. The following GSECs are then added to the model to ensure that each customer served by vehicle $r$ belongs to a path that is connected to the depots and does not form a cycle with other vertices of $S$. 
\begin{align}
\sum_{(u,v) \in \delta(S)} x_{uvr} \geq 2 y_{ir}, \forall S \subset V, \{d,a\} \subseteq S, \forall i \in V \setminus S, \forall r \in F \label{eq:gsec1}
\end{align}

We also add two categories of constraints, which are detailed below and are equivalent to the GSECs, to strengthen the model.

\begin{align}
\sum_{(u,v) \in \gamma(S)} x_{uvr} & \leq \sum_{i\in S \setminus \{d, a\}} y_{ir} - y_{jr} + 1, \forall S \subset V, \{d,a\} \subseteq S, \forall j \in V \setminus S, \forall r \in F \label{eq:gsec21} \\
\sum_{(u,v) \in \gamma(U)} x_{uvr} & \leq \sum_{i\in U} y_{ir} - y_{jr}, \forall U \subseteq V^{-}, \forall j \in U, \forall r \in F \label{eq:gsec22}
\end{align}

On the other hand, our approach requires a check on the absence of subtours for an optimal solution of the current incomplete model (i.e. while relaxing constraints \eqref{miptop:ctsubtour}), so that the global optimality can be claimed. In our model, each \emph{strong connected component} of the subgraph associated with a tour of the solution represents a subtour, thus the checking can be done by examining the corresponding subgraphs. This will be detailed in Section \ref{ssec:CuttingPlane}.

\section{Efficient cuts}\label{sec:domprop}
Reduction of the search space is often desired in solving a MIP. This can be done by either removing irrelevant components from the linear formulation, e.g. those that certainly does not belong to any optimal solution, or by favoring some special structures inside the optimal solutions, e.g. reduction of the symmetry.
The cuts that we added to our basic problem include some dominance properties as symmetry breaking inequalities, boundaries on profits and number of served customers, cuts that enforces mandatory customers and cuts that remove inaccessible customers and arcs. Moreover, some additional cuts are based on the clique and the independent sets deduced from the incompatibilities between solution components.

\subsection{Symmetry breaking cuts}
Tours of the optimal solutions can be sorted according to a specific criterion, i.e. the amount of collected profits, the number of customers or the tour length. Based on the experimental report in \citep{bib:Dang13a}, we focus exclusively on solutions in which the profits of tours are in ascending order. The following constraints are added to the model to ensure the symmetric breaking on profits.

\begin{equation}
\sum_{i \in V^{-}} y_{i(r+1)} p_i - \sum_{i \in V^{-}} y_{ir} p_i \leq 0, \forall r \in F \setminus \{m\} \label{eq:symbrk}
\end{equation}

Without these constraints, for each feasible solution having different profits among its tours, there are at least $(m!-1)$ equivalently feasible solutions. Adding these constraints will remove these equivalent solutions from the search space and only retain the one having the profits of its tours in ascending order. Thus the size of the search space can be largely reduced.

\subsection{Irrelevant components cuts}

One simple way of reducing the size of the problem is to deal only with accessible customers and arcs. A customer is considered as \emph{inaccessible} if by serving only that customer, the travel cost of the resulting tour exceeds the cost limit $L$. In a similar way, we detect an \emph{inaccessible} arc when the length of the tour directly connecting the depots to that arc exceeds $L$. To make a proper linear formulation, all \emph{inaccessible} customers and arcs are eliminated at the beginning from the model by adding the following constraints. Here $i$ is an inaccessible customer (resp. $(i,j)$ is an inaccessible arc).
\begin{align}
\sum_{r \in F} y_{ir} = 0 \\
\sum_{r \in F} x_{ijr} = 0
\end{align}

\subsection{Boundaries on profits and numbers of customers served}\label{sec:ublbprofit}
\cite{bib:Dang13a} proposed in their paper a set of efficient dominance properties that aims to reduce the search space by bounding the characteristics of each tour or subset of tours. The idea is to solve within a limited time budget, instances derived from the original problem to gain useful information for the construction of the added cuts. The derived instances are often smaller than the original one and hopefully easier to solve, or at least to bound.

Before going in the details of these properties, we must clarify some notation. For each instance $X$ with $m$ vehicles, define $X_{I}$ to be the modified instance for which the profits of each customer is set to $1$ instead. We also use $X^{g}$ to denote the modified instance $X$ by reducing the number of available vehicles to $g$ ($g \leq m$). For $g=m$, we have the original instance $X$. Note that the two modifications can be applied at the same time, in this case instance $X_{I}^{g}$ is obtained. Finally, we denote by $\LB(X)$ (resp. $\UB(X)$) a lower (resp. an upper) bound of an arbitrary instance $X$. The following valid inequalities are added to the model to restrict the profits that each tour or subset of tours can have.
\begin{align}
\sum_{r \in H} \sum_{i \in V^{-}} y_{ir} p_i & \leq \UB(X^{|H|}), \forall H \subset F \label{eq:ubp}\\
\sum_{r \in H} \sum_{i \in V^{-}} y_{ir} p_i + \UB(X^{m-|H|}) & \geq \LB(X), \forall H \subseteq F \label{eq:lbp}
\end{align}

Inequalities \eqref{eq:ubp} are trivial since the sum of profits of any $|H|$ tours on the left-hand side cannot exceed the optimal profit of the instance with exactly $|H|$ vehicles or at least an upper bound of this instance, i.e. the right-hand side.

Inequalities \eqref{eq:lbp} work in the opposite direction by applying a lower bound to the profit of each tour and each subset of tours. The inequalities might appear to be redundant with the objective of optimization. However when applied to subsets of tours, the constraints will eliminate unbalanced solutions, e.g. the one with one tour having many customers and the other tours being almost empty, from the search space.

In the same fashion as \eqref{eq:ubp}, the numbers of customers per tour or per subset of tours are bounded from above using inequalities \eqref{eq:ubc}. On the other hand, it is more difficult to bound these numbers from below since their values do not necessarily correlate with the objective value of TOP. A modification of the model (rather than a simple modification of the instance) is performed in order to determine a lower bound for the number of customers of each tour. This modification is done as follows. We consider the modified instance, denoted by $\bar{X}^1_I$, where the objective function is reversed to minimization, i.e. minimizing the number of served customers, while satisfying both constraints \eqref{eq:ubp} and \eqref{eq:lbp} for $|H|=1$. Solving this instance provides the value of $\LB(\bar{X}^1_I)$, which enables us to lower bound the number of customers of each tour of $X$. The following valid inequalities are then added to the model to restrict the number of customers served in each tour or subset of tours.

\begin{align}
\sum_{r \in H} \sum_{i \in V^{-}} y_{ir} & \leq \UB(X^{|H|}_I), \forall H \subset F \label{eq:ubc} \\
\sum_{i \in V^{-}} y_{ir} & \geq \LB(\bar{X}^1_I), \forall r \in F \label{eq:lbc}
\end{align}

In implementation, inequalities \eqref{eq:ubp} - \eqref{eq:lbc} are applied similarly to dynamic programming, as follows. The required values of $\LB$ and $\UB$ are first computed for the instance with $|H|=1$ , then the obtained values are used in the cuts to solve the other instances ($|H| \leq m$). We recall that inequalities \eqref{eq:lbc} are limited to a single tour and not subsets of tours. Since the value of $\UB(X^{m-1})$ is needed for the model of $\bar{X}^1_I$, $\LB(\bar{X}^1_I)$ can only be computed after solving all the other subproblems (or derived instances).

\subsection{Mandatory customers cuts}\label{sec:mandatory}

Given an instance $X$ of TOP, a high quality $\LB(X)$ can often be computed efficiently with heuristics. Therefore, it could be possible to locate a set of customers of $X$, the so-called \emph{mandatory} ones, for which without one of those customers a solution with the objective value at least as large as $\LB(X)$ cannot be achieved. 

The formal definition is the following. Here we use $X\setminus\{i\}$ to designate the modified instance $X$ with customer $i$ removed.

\begin{defn}\label{def:mandatory}
A customer $i$ of $X$ is \emph{mandatory} if $\UB(X\setminus\{i\})<\LB(X)$. 
\end{defn}

Once identified, mandatory customers have to be all served in an optimal solution. The following cuts can then be added to enforce the presence of a mandatory customer $i$ in $X$.
\begin{align}
\sum_{r \in F} y_{ir} = 1 \label{eq:mandatory}
\end{align}

\subsection{Valid inequalities based on incompatibilities}\label{sec:cliquecut}

If two given customers are too far away from each other because of the travel length/cost limitation, then it is unlikely that they can be served by the same vehicle. This observation leads us to the concept of \emph{incompatibility} between customers, from which additional inequalities can be deduced \citep{bib:Manerba15, bib:Gendreau16}. Moreover, the idea can also be generalized to other pairs of components of the problem, i.e. {\it customer-tour}, {\it customer-arc} or {\it arc-arc}. In this work, we focus on the two incompatibilities: between customers and between arcs.

\subsubsection{Incompatibility graphs}\label{ssec:incomp}

Given two customers $i$ and $j$ of instance $X$, we use $X\cup\{[i \sim j]\}$ to denote the modified instance/model where enforcing the two customers $i$ and $j$ to be served by the same vehicle is imposed as a constraint. Similarly, $X\cup\{[(u,v) \sim (w,s)]\}$ denotes the modified instance/model in which arcs $(u,v)$ and $(w,s)$ are imposed to be used by the same vehicle. The two graphs of incompatibilities are formally defined as follows.

\begin{defn}\label{def:ginc}
Given an instance $X$ of the TOP modelled by the directed completed graph $G=(V,A)$, the graph of incompatibilities between customers is $G^{Inc}_{V^{-}}=(V^{-}, E^{Inc}_{V{-}})$ and between arcs is $G^{Inc}_{A}=(A, E^{Inc}_{A})$ where	
\begin{align*}
E^{Inc}_{V^{-}} &= \{[i,j] \mid i, j \in V^{-}, \UB(X\cup\{[i \sim j]\}) < \LB(X)\}, \\
E^{Inc}_{A} &= \{[i,j] \mid i = (u,v), j = (w,s) \in A, \UB(X\cup\{[(u,v) \sim (w,s)]\}) < \LB(X)\}.
\end{align*}
\end{defn}

In other words, two components are \emph{incompatible} if they do not appear in the same tour of any optimal solution of instance $X$. In general, it is difficult to fully construct the two graphs of incompatibilities. However, they can be initialized
as follows. Here, $\MinLen(S)$ denotes the length of the shortest path from $d$ to $a$ and containing all vertices (or all arcs) of $S \subseteq V^{-}$ (or $\subseteq A$).

\begin{prop}\label{prop:ginit}
Let $G=(V,A)$ be the model graph of instance $X$, it holds that
\begin{align*}
  &\{[i,j] \mid i \in V^{-}, j \in V^{-}, \MinLen(\{i, j\})>L\} \subseteq E^{Inc}_{V^{-}}\text{, and}  \\
  &\{[i,j] \mid i = (u,v) \in A, j = (w,s) \in A, \MinLen(\{(u, v),(w,s)\})>L\} \subseteq E^{Inc}_{A}.
\end{align*}

\end{prop}

Of course, once initialized the graphs can be filled with more edges using Definition \ref{def:ginc}. The density of the becoming graphs will depend on the computation of $\UB$ and $\LB$. We can use the following linear program, combining with other cuts we have developed, to compute the required $\UB$.

\begin{prop}\label{prop:force}
Let $X$ be an instance $X$ of TOP and $i, j$ be its two customers, the linear model of $X\cup\{[i \sim j]\}$ is obtained by adding to that of $X$ the following constraints:
\begin{align}
\sum_{r \in F} y_{ir} &= \sum_{r \in F} y_{jr} = 1 \label{eq:fc1}\\
y_{ir} &=  y_{jr}, \forall r \in F  \label{eq:fc2}
\end{align}
Similarly, adding the following constraints to the linear program of $X$ will model  $X\cup\{[(u,v) \sim (w,s)]\}$.
\begin{align}
\sum_{r \in F} x_{uvr} &= \sum_{r \in F} x_{wsr} = 1 \label{eq:fa1} \\
x_{uvr} &= x_{wsr}, \forall r \in F  \label{eq:fa2}
\end{align}
\end{prop}

\subsubsection{Clique cuts}\label{ssec:clique}
A clique in an undirected graph is a subset of vertices that are pairwise adjacent. Thus, serving a customer (or using an arc) belonging to a clique of $G^{Inc}_{V^{-}}$ (or $G^{Inc}_{A}$)  by a vehicle will exclude all other customers (or arcs) of the clique from being served by the same vehicle. Therefore, each vehicle can only serve (or use) at most one element of the clique. Based on this observation, the following cuts hold for $G^{Inc}_{V^{-}}$ and $G^{Inc}_{A}$, with $K$ (resp. $Q$) represents a clique of $G^{Inc}_{V^{-}}$ (resp. $G^{Inc}_{A}$).
\begin{align}
\sum_{i \in K} y_{ir} & \leq 1, \forall r \in F \label{eq:clvcut} \\
\sum_{[u,v] \in Q} x_{uvr} & \leq 1, \forall r \in F \label{eq:clecut}
\end{align}

A clique is \emph{maximal} if it cannot be extended to a bigger one by adding more vertices, and a maximal clique is \emph{maximum} if it has the largest cardinality over the whole graph. Large and maximal cliques are preferred in inequalities \eqref{eq:clvcut} and \eqref{eq:clecut} since they provide tighter formulations. The difficulty is that the number of maximal cliques in a general graph is exponential in terms of the number of vertices and finding the maximum clique is an NP-Hard problem \citep{bib:garey79}. However, efficient methods to find those cliques or subset of them exist in the literature and work very well in our graphs. The details are discussed in Section \ref{sec:CuttingPlanegc}.

\subsubsection{Independent set cuts}\label{ssec:iss}
As opposed to a clique, an \emph{independent set} is a set of vertices in a graph such that no two of which are adjacent. In that case, the vertices are also called pairwise independent. Maximal and maximum independent sets are defined in the same way as for cliques, e.g. adding any vertex to a maximal independent set will invalid the independences between the vertices of the set, and a maximum independent set is one of the largest sets among the maximal ones.

The independent-set cuts are based on the following idea. Let us consider $G^{Inc}_{V^-}$ as an example of graph and let $S$ be a subset of $V^-$, we define $\alpha_S$ to be the size of a maximum independent set of $G^{Inc}_{V^-}(S)$, the subgraph vertex-induced by $S$. It is clear that no more than $\alpha_S$ components of $S$ can be served in the same tour, e.g. $\sum_{i \in S} y_{ir} \leq \alpha_S$ is a valid cut for any tour $r$. Furthermore, if we consider $S$ to be the set of neighbor vertices of a vertex $i$ in $G$ denoted by $N_i$, then we can add the cut $\alpha_i y_{ir} + \sum_{j \in N_j} y_{jr} \leq \alpha_i$
(here $\alpha_i$ is a short notation for $\alpha_{N_i})$. This particular cut embeds the relationship between $i$ and $N_i$, plus the information on the maximum independent set of $N_i$. The same idea can be generalized to $G^{Inc}_{A}$, where we denote by $N_{ij}$ the set of neighbor arcs of an arc $(i, j)$ in $G^{Inc}_{A}$, and the following inequalities summarize the valid cuts.
\begin{align}
\alpha_i y_{ir} + \sum_{j\in N_i} y_{jr} & \leq \alpha_i, \forall i\in V^-, \forall r \in F \label{eq:advclvcut}\\
\alpha_{ij} x_{ijr} + \sum_{(u,v)\in N_{ij}} x_{uvr} & \leq \alpha_{ij}, \forall (i,j)\in A, \forall r \in F \label{eq:advclecut}
\end{align}

Finding a maximum clique is NP-Hard and so is to find a maximum independent set \citep{bib:garey79}. However, the above inequalities also hold for $\alpha$ being an upper bound of the size of a maximum independent set. The following principle allows us to approximate such an upper bound. Recall that a partition of vertices of a graph into disjoint independent sets is a coloring of the graph, e.g. each independent set is assigned to a color. It is well-known that the number of colors used in any such coloring is an upper bound of the size of a maximum clique of the graph. From the perspective of the complementary graph, any partition of the vertices into disjoint cliques provides an upper bound on the size of a maximum independent set. Again, efficient algorithm to find large cliques can be used to make such a partition and then to compute the upper bound of $\alpha_i$. This procedure is detailed in the next section.

\section{Cutting-plane and global scheme}\label{sec:CuttingPlanegc}

In this section, our global Cutting-Plane Algorithm (CPA) is first described to show the different operations performed to reach the best solution. Some supplementary information is required for its execution, particularly for the construction of the efficient cuts. These computations are detailed in the Constraint-Enhancement algorithm (CEA).

\subsection{Cutting-Plane algorithm}\label{ssec:CuttingPlane}

Our global algorithm is a cutting-plane one. However, we also use it to solve intermediate models with fewer numbers of vehicles (and sometimes with modified constraints/objectives). In our implementation, we only focus on the elimination of subtours and on the refinement of the search space using the developed cuts, while the other aspects of the resolution, e.g. the branch-and-cut in solving the integer program, are left for the MIP solver. This is similar to the approach of \citep{bib:Pferschy13} which was developed in the context of TSP. The steps of our CPA are as follows. 

At first, the basic model is built using constraints \eqref{miptop:ctvisit}-\eqref{miptop:ctlength} and \eqref{miptop:ctinteger} with the objective function \eqref{miptop:obj} and some initial cuts. Indeed, some pre-computations are performed beforehand to gain useful information for the initial cuts. Only a small time budget is allowed for these pre-computations, however this can lead to a significant strengthening of the model later on.

During the pre-computation phase, the irrelevant components of $X$, i.e. \emph{inaccessible} customers and arcs, are first detected and removed from the model. Then the graphs of incompatibilities between customers and arcs are initialized, and some early cliques and independent sets are extracted from them using the metaheuristic described in \citet{bib:Dang12}. Based on these sets, the associated clique and independent set cuts are formulated and added to the model. Finally the symmetry breaking cuts are added and the solving procedure begins. A feasible solution is generated using a heuristic of \citet{bib:Dang13b} and provided to the MIP solver as a starting solution.

Before going in the main loop of the solving process, the MIP solver is setup with some branching rules. In TOP, the objective function aims to maximize the collected profits from the visited customers, therefore, selecting the correct customers from the beginning appears to be crucial. Thus, our branching rules prioritize $y_{ir}$ first then $x_{ijr}$ \citep{bib:Boussier07, bib:Aragao10}.

\begin{algorithm}[!h]
	\caption{Cutting-Plane algorithm (CPA).}\label{alg:cuttingPlaneAlgorithm}
	\KwIn{Instance $X$, cuts $\D(X)$, timer $\TM$, indicator $\VarOriginal$}
	\KwOut{Bound $\UB(X)$, solution $\SOL(X)$, indicator $\Opt(X)$}
	
	\Begin{
		$\VarStep \leftarrow 1$\;
		$\Opt(X) \leftarrow$ \textbf{false}\;
		$\MIPS \leftarrow$ create new MIP Solver\;
		$\UB(X) \leftarrow$ sum of profits of all customers of $X$\;
		$\SOL(X) \leftarrow$ a feasible solution of $X$ \citep[see][]{bib:Dang13b}\;
		$\LB(X) \leftarrow \PFT(\SOL(X))$\;    	
		MIPS.model($X$, $\D(X)$) (see Sections \ref{sec:lp} and \ref{sec:onsidealg})\;
		MIPS.initialize($\SOL(X)$)\;
		\Repeat{{\upshape ($\Opt(X)=${\bf true}) or ($\TM$.expired())}} {
			$\{\UB, \SOL, \Opt\} \leftarrow$ MIPS.solve($\TM$)\;
			\lIf{\upshape ($\UB<\UB(X)$)}{$\UB(X) \leftarrow \UB$}
			\If{\upshape ($\Opt=${\bf true})}{
				\lIf{\upshape ($\PFT(\SOL)<\UB(X)$)}{ $\UB(X) \leftarrow \PFT(\SOL)$}
				$\{T_r\}_{r\in F} \leftarrow$ extract subtours from SOL\;
				$\{S_r\}_{r\in F} \leftarrow$ extract tours from SOL\;
				\If{\upshape ($\PFT(\bigcup_{r \in F} S_r)>\LB(X)$)}{ 
					$\SOL(X) \leftarrow \{S_r\}_{r \in F}$\;
					$\LB(X) \leftarrow \PFT(\SOL(X))$\;
				}
				\eIf{\upshape ($|\bigcup_{r\in F} T_r|=0$) or ($\LB(X)=\UB(X)$)}{
					Opt($X$) $\leftarrow$ {\bf true}\;
				} {
				MIPS.add(GSEC($\{T_r\}_{r\in F}$)) (see Section \ref{sec:lp})\;
				(add clique cuts, see Section \ref{ssec:clique})
				MIPS.add(FindCliques($G^{Inc}_{V^{-}}[\bigcup_{r\in F}(T_r \cup S_r)]$)) \;
				MIPS.add(FindCliques($G^{Inc}_{A}[\bigcup_{r\in F}(T_r \cup S_r)]$))\;
				\If{\upshape ($\VarOriginal=${\bf true})}{D(X)$\leftarrow$ CEA($X$, $\LB(X)$, Step)\;
					MIPS.add($\D(X)$)\;
					$\VarStep \leftarrow \VarStep + 1$;
				}
			}
		}
	}
}
\end{algorithm}

Algorithm \ref{alg:cuttingPlaneAlgorithm} summarizes the remaining steps of our CPA. In each iteration of the main loop, the MIP solver is called to solve the linear model and an integer solution is obtained. Tarjan's algorithm \citep{bib:tarjan72} is then applied on this solution to check if it contains any subtour. Recall that a directed graph is strongly connected if for any given pair of vertices there exist paths linking them in both directions. A strong connected component of a directed graph is a subset of its vertices such that the induced subgraph is strongly connected and that the subset cannot be extended by adding more vertices. Since in our formulation, the graph is directed and the depots are separated vertices, the vertices of a subtour can only belong to a strong connected component of the subgraph. That is to say the total absence of those components for each subgraph, which can be polynomially detected \citep{bib:tarjan72}, implies the global optimality and the CPA is terminated by returning the solution. Otherwise, the solution is \emph{suboptimal}. The associated constraints \eqref{eq:gsec1}, \eqref{eq:gsec21} and \eqref{eq:gsec22}, deduced from the suboptimal solution, are then added to the linear model to eliminate the subtours. Furthermore, the subgraphs of $G^{Inc}_{V^{-}}$ and $G^{Inc}_A$, which are associated to the vertices and arcs of the suboptimal solution, are extracted. Some maximal cliques are then generated from those subgraphs, and the corresponding constraints \eqref{eq:clvcut} and \eqref{eq:clecut} are added to the linear model. Next, if we are solving the original problem (indicated by the boolean $\VarOriginal$), the CEA is called to generate a set of efficient constraints for the model. This algorithm is described in Section~\ref{sec:onsidealg}. Once all the cuts are added to the model, the CPA goes to the next iteration where the same solving process is repeated (with the modified model). On the event that the predefined time limit (indicated by the timer $\TM$) is run out, the algorithm is terminated and the best bound computed so far is returned for the instance/model.

Algorithm \ref{alg:cuttingPlaneAlgorithm} takes as inputs an instance $X$, a set of cuts $\D(X)$, and a boolean indicator $\VarOriginal$. It also requires a mixed integer programming solver and a timer to operate. The algorithm returns an upper bound $\UB(X)$, a feasible solution $\SOL(X)$ and a boolean indicator $\Opt(X)$ telling the optimality of $\SOL(X)$ before the expiration of the timer. For the purpose of simplification, tours of the initially generated solutions are supposed to be sorted to match inequalities \eqref{eq:symbrk}. We also assume that the mixed integer programming solver can be adapted to support the following operations: \emph{model} to construct the linear integer model based on $X$ and D$(X)$ and according to our specification, including branching rules; \emph{initialize} to provide a feasible starting solution to the solver; \emph{add} to complete the model with efficient cuts; and finally \emph{solve} to try to solve the model until the expiration of a timer. The output of \emph{solve} is similar to Algorithm \ref{alg:cuttingPlaneAlgorithm}: a scalar reporting an upper bound, a feasible solution (which can be empty) and a boolean reporting the optimality.

\subsection{Generation of efficient cuts}\label{sec:onsidealg}

To solve an original instance of TOP, our CPA needs strong constrained models in its earlier iterations. For this purpose, CEA is called and the counter $\VarStep$ of the main algorithm is passed to it as a parameter. For each value of $\VarStep$ less than $m + 1$, only one type of cuts is computed and the produced cuts are added to the model. The details of the procedure are given in Algorithm \ref{alg:CEAlgorithm}. Note that with the efficient constraints along the way, some easy instances can be solved in less than $m+1$ iterations.

\begin{algorithm}[!h]
	\caption{Constraint-Enhancement algorithm (CEA).}\label{alg:CEAlgorithm}
	\KwIn{Instance $X$, bound $\LB(X)$, integer $\VarStep$}
	\KwOut{Cuts D$(X)$}
	\Begin{
		\If{\upshape $\VarStep \leq m-1$}{
			(solve intermediate models, see Section \ref{sec:ublbprofit})\\
			$\{\UB, \SOL, \Opt\} \leftarrow$ CPA($X^{\VarStep}$, $\D(X)$, $\TM_1$, {\bf false})\;
			D($X$) $\leftarrow$ update from $\{\UB, \SOL, \Opt\}$\;
			$\{\UB, \SOL, \Opt\} \leftarrow$ CPA($X^{\VarStep}_I$, $\D(X)$), $\TM_1$, {\bf false})\;
			D($X$) $\leftarrow$ update from $\{\UB, \SOL, \Opt\}$\;
			\If{\upshape ($\VarStep=m-1$)}{
				$\MIPS \leftarrow$ create new MIP Solver\;
				MIPS.model($\bar{X}^{1}_I$, $\D(X)$)\;
				$\{\UB, \SOL, \Opt\} \leftarrow$ MIPS.solve($\TM_1$)\;
				D($X$) $\leftarrow$ update from $\{\UB, \SOL, \Opt\}$\;			
			}
		}
		\If{\upshape $\VarStep = m$}{
			(identify mandatory customers, see Section \ref{sec:mandatory})\\
			$\VarMandatory \leftarrow \emptyset$\;  
			\ForEach{$i \in V^-$}{
				$\{\UB, \SOL, \Opt\} \leftarrow$ CPA($X\setminus\{i\}$, $\D(X)$, $\TM_1$, {\bf false})\;
				\If{$\UB<\LB(X)$}{
					$\VarMandatory \leftarrow \VarMandatory \cup \{i\}$\;
				}
			}
			D($X$) $\leftarrow$ update with $\VarMandatory$ as mandatory customers\;
		}
		\If{\upshape $\VarStep = m+1$}{
			(enhance incompatibilities, see Section \ref{ssec:incomp})\\
			\ForEach{$(i,j) \in A$}{
				$\{\UB, \SOL, \Opt\} \leftarrow$ CPA($X\cup\{i \sim j\}$, $\D(X)$), $\TM_1$, {\bf false})\;
				\If{$\UB<\LB(X)$}{
					update $G^{Inc}_{V^{-}}$\;
				}
				\For{$(u,v) \in A$}{
					$\{\UB, \SOL, \Opt\} \leftarrow$ CPA($X\cup\{[(i,j) \sim (u,v)]\}$, $\D(X)$), $\TM_1$, {\bf false})\;
					\If{$\UB<\LB(X)$}{
						update $G^{Inc}_{A}$\;
					}
				}
			}
			(identify clique/independant-set cuts, see Section \ref{ssec:clique}, \ref{ssec:iss})\\
			D($X$) $\leftarrow$ update from FindCliques($G^{Inc}_{V^{-}}$),  FindCliques($G^{Inc}_{A}$)\;
		}
	}
\end{algorithm}

The first type of cuts to be generated is the one corresponding to the boundaries on profits and numbers of customers for each subset of tours. For each subproblem with the number of vehicles being reduced to $\VarStep$ ($\VarStep\leq m - 1$), upper bounds for the feasible profit and the feasible number of customers are computed using the same CPA as described in the previous section (except that $\VarOriginal$ is set to false). The corresponding constraints are then generated and added to $\D(X)$, the storage of all additional information and cuts. In the case of $\VarStep$ equal to $m - 1$, before returning to the main algorithm, a lower bound on the feasible number of customers for a single vehicle is calculated. This calculation makes use of the information accumulated in $\D(X)$ and expands it further with the obtained lower bound.

When the main algorithm reaches iteration $m$, efficient constraints of the second type is constructed, and mandatory customers are located to strengthen the model. These customers are identified based on Definition \ref{def:mandatory}: the required $\LB$ is computed using a constructive heuristic from \citep{bib:Dang13b} while the required $\UB$ is computed with our CPA, but now formulated for the instances $X \setminus \{i\}$. Once a mandatory customer is located, it is immediately added to $\D(X)$ so that the information can be used in the subsequent iterations.

Being constructed at iteration $m + 1$ of the main algorithm, clique and independent-set cuts are the third type of cuts. First, graphs $G^{Inc}_{V^{-}}$ and $G^{Inc}_{A}$ are initialized with Property \ref{prop:ginit}. Since the verification of $\MinLen(\cdot)$ in this case maximally involves $4$ customers, a complete enumeration is inexpensive and manageable. In addition, these initial graphs can be computed beforehand and stored for each instance. The graphs are then made more dense using their definition: lower bounds $\LB(X)$ are due to the results of \citep{bib:Dang13b}, and $\UB(X\cup\{[i\sim j]\})$ and $\UB(X\cup\{[(i,j)\sim(u,v)]\})$ are computed with the CPA, while adding constraints \eqref{eq:fc1}-\eqref{eq:fa2} to construct the desired models. Next, the clique cuts and independent set cuts are generated from $G^{Inc}_{V^{-}}$ and $G^{Inc}_{A}$ and used as general constraints. For each vertex in the associate incompatibility graph, we determine a large maximal clique containing the vertex using the metaheuristic from \citep{bib:Dang12}. On the other hand, using the very same heuristic algorithm, a partition of each $N_i$ (resp. $N_{ij}$) into disjoint cliques can be constructed. For example, first find a large clique, then remove its vertices from the graph and continue finding cliques on the remaining graph. Thus, upper bounds for $\alpha_i$ (resp. $\alpha_{ij}$) are computed.

We note that to generate the three types of efficient cuts, the CPA is called with a time limit configured by timer $\TM_1$.

\section{Numerical results}\label{sec:numresult}

Our algorithm is coded in C++. Experiments were conducted on an AMD Opteron $2.60$ GHz and CPLEX $12.5$ was used as MIP solver. We used the same two-hours limit of solving time as in \citet{bib:Boussier07, bib:Aragao10}, of which at most a one-hour limit is given to generate all the efficient cuts. This one hour limit is divided between solving the smaller problems, locating the mandatory customers and extending the incompatibility graphs. We first evaluated the usefulness of the proposed components by activating each type of the efficient cuts without the other types, then by activating all of them together.

\subsection{Benchmark instances}
We evaluated our approach on a set of TOP instances proposed by \citet{bib:Chao96a}. This benchmark comprises $387$ instances and is divided into $7$ data sets. In each data set, the positions and the profits of the customers are identical for all instances. However, the number of vehicles varies from $2$ to $4$ and the travel length limit $L$ is also different between instances. The latter causes a variation of the number of accessible customers (denoted by~$n'$) even when the number of vehicles is fixed. Each instance is named according to the data set to which it belongs, the number of available vehicles and a letter that designates the travel length $L$. However, note that an identical letter inside a data set does not necessarily imply the same value of $L$ when the number of vehicles changes. The characteristics of each data set are reported in Table \ref{tab:instances}.

\begin{table}[h]
\vspace{10pt}
\caption{Instances of \citet{bib:Chao96a}.}\label{tab:instances} 
\begin{center}
\begin{tabular}{cccccccc}
\toprule
Set&1&2&3&4&5&6&7\\
\midrule
\#{}Inst. &$54$&$33$&$60$&$60$&$78$&$42$&$60$\\
$n$       &$30$&$19$&$31$&$98$&$64$&$62$&$100$\\
$n'$      &0-30&1-18&2-31&0-98&0-64&0-62&0-100\\
$m$       &$2$-$4$&$2$-$4$&$2$-$4$&$2$-$4$&$2$-$4$&$2$-$4$&$2$-$4$\\
$L$       &$3.8$-$22.5$&$1.2$-$42.5$&$3.8$-$55$&$3.8$-$40$&$1.2$-$65$&$5$-$200$&$12.5$-$120$\\
\bottomrule
\end{tabular}
\end{center}
\end{table}

\subsection{Component evaluation}
We present in Table \ref{tab:impact} the results obtained with the basic model, then those obtained while separately applying the GSECs, the dominance properties (see Section~\ref{sec:domprop}) and the valid inequalities (see Section~\ref{sec:cliquecut}). The last main column shows the results of the global algorithm by activating all of the components together. In this table, columns \#Opt, $CPU_{avg}$ and Gap respectively represent, for each set, the number of instances being solved to optimality, the average computational time in seconds on the subset of common instances being solved by all the configurations and the average percentage gap. Note that the percentage gap of an instance is calculated as follows: $\mathrm{Gap} = 100 \times \frac{\UB - \LB}{\UB}$, where $\UB$ and $\LB$ are the upper and lower bounds computed for the instance.

\afterpage{
{\setlength{\tabcolsep}{2.5pt
\begin{landscape}
\begin{table}[h!]
\vspace{0.7cm}
\caption{Impact of the proposed cuts.}
\label{tab:impact} 
\begin{center}
\begin{tabular}{c>{\columncolor[RGB]{235,235,235}}cccc>{\columncolor[RGB]{235,235,235}}cccc>{\columncolor[RGB]{235,235,235}}cccc>{\columncolor[RGB]{235,235,235}}cccc>{\columncolor[RGB]{235,235,235}}cccc}
\toprule
\multirow{2}{*}{Set}&&\multicolumn{3}{c}{Basic model}&&\multicolumn{3}{c}{GSECs}&&\multicolumn{3}{c}{Dominance properties}&&\multicolumn{3}{c}{Valid inequalities}&&\multicolumn{3}{c}{All cuts}\\
 & &\#Opt&$CPU_{avg}$&Gap&&\#Opt&$CPU_{avg}$&Gap&&\#Opt&$CPU_{avg}$&Gap&&\#Opt&$CPU_{avg}$&Gap&&\#Opt&$CPU_{avg}$&Gap\\
\midrule
$1$&&$35/54$&$496.5$&$5.04$&&$53/54$&$13.2$&$0.54$&&$54/54$&$5.3$&$0$&&$54/54$&$2.9$&$0$&&$54/54$&$1.7$&$0$\\
$2$&&$33/33$&$5.5$&$0$&&$33/33$&$1.8$&$0$&&$33/33$&$0.6$&$0$&&$33/33$&$0.1$&$0$&&$33/33$&$0.03$&$0$\\
$3$&&$42/60$&$599.9$&$3.41$&&$55/60$&$150.9$&$0.25$&&$58/60$&$26.9$&$0.5$&&$60/60$&$10.3$&$0$&&$60/60$&$6.24$&$0$\\
$4$&&$23/60$&$323.5$&$3.19$&&$17/60$&$390.5$&$4.28$&&$22/60$&$200.4$&$2.05$&&$23/60$&$81$&$2.31$&&$30/60$&$66.6$&$0.01$\\
$5$&&$23/78$&$318$&$12.9$&&$24/78$&$40.2$&$21.11$&&$37/78$&$5.4$&$6.35$&&$36/78$&$2.6$&$6.65$&&$54/78$&$0.95$&$0.01$\\
$6$&&$33/42$&$48.7$&$0.4$&&$33/42$&$63.8$&$3.53$&&$41/42$&$4.5$&$0.11$&&$39/42$&$3.8$&$0.76$&&$42/42$&$1.9$&$0$\\
$7$&&$14/60$&$46.3$&$12.88$&&$18/60$&$3.5$&$13.08$&&$22/60$&$2.1$&$7.24$&&$24/60$&$1.3$&$5.75$&&$27/60$&$0.28$&$0.03$\\
\midrule
Total&&$204/387$&$294.2$&$7.0$&&$233/387$&$80.4$&$7.4$&&$267/387$&$26.7$&$2.8$&&$269/387$&$11.24$&$2.6$&&$300/387$&$8.21$&$0.01$\\
\bottomrule
\end{tabular}
\end{center}
\end{table}
\end{landscape}
}
}
}

Compared to the results obtained in the basic model, all the proposed components independently and positively affect the outcomes of the algorithm. As shown in Table \ref{tab:impact}, GSECs largely help increase the numbers of instances being solved except for some instances from the large sets, where a significant number of GSECs should be added to start having some progress in the resolution. The valid inequalities, which include the clique and the independent-set cuts, mainly contribute to the reduction of the computational times and the average gaps. The dominance properties, which comprise the symmetry breaking, mandatory customers, irrelevant component and boundaries on profits and number of customers, have an effect similar to that of the valid inequalities, specially on the numbers of instances being solved to the optimality and the average gaps. The relatively large computational times obtained while applying the dominance properties are due to the amounts of time spent on solving subproblems.

On the other hand, we notice from the last column of Table \ref{tab:impact} that applying all the proposed components together remarkably improves the number of instances being solved, reaching $300$ of the $387$ instances. This also implies a reduction of the average gaps between the upper and the lower bounds. In addition, the average computational time of the global algorithm decreased from $294.2$s with the basic model to $8.21$s with all the enhanced components applied.

\subsection{Comparison with other exact methods in the literature}

We first compare our proposed method with the other exact methods in the literature on a per-instance basis. Since \citet{bib:Aragao10} did not report the detailed results of their algorithm, we restricted our comparison to the results of the B-P algorithm of \citet{bib:Boussier07}, the B-C algorithm of \citet{bib:Dang13a} and the B-P-2R and the B-C-P algorithms of \citet{bib:Keshtkarana15}. The computational experiments of B-P were carried out on a Pentium IV, $3.2$ GHz while those of B-C on an AMD Opteron, $2.60$ GHz and those of B-P-2R and B-C-P on a single core of an Intel Core i7 $3.6$ GHz. 

Table \ref{tab:comparison_literature} reports the results of the instances which are solved by at least by one of the five methods (but not by all of them). In this table, columns $Instance$, $n$, $m$, and $L$ respectively show the name of the instance, the number of accessible customers, the number of vehicles and the travel cost limit. Columns $UB$, $LB$, and $CPU$ report respectively the upper bound, lower bound and computational time in seconds for each method and for each instance when available. For B-P \citep[see][]{bib:Boussier07}, the reported $CPU$ is the time spent on solving both the master problem and the subproblems until the optimality is proven. For B-C \citep[see][]{bib:Dang13a}, the $CPU$ time includes the computational times for both, the presolving and solving phases. For B-P-2R and B-C-P \citep[see][]{bib:Keshtkarana15}, the $CPU$ time is reported for the whole solving process. In our method, we consider the $CPU$ time as the time spent in the global algorithm with the required computational time to generate the efficient cuts. For some instances, dashes ``$-$'' are used in $UB$ and $LB$ columns when the corresponding values were not found and tildes ``$\sim$'' are used in $CPU$ column to show that the optimalities were not proven within $7200s$ of the time limit.

\begin{small}
{\setlength{\tabcolsep}{2.5pt}
\begin{landscape}
\begin{longtable}{ccccccccccccccccccc}
	\caption{Comparison between our results and the literature on the standard benchmark.}\\
	\toprule
	\multirow{2}{*}{$Instance$}&
	\multirow{2}{*}{$n$}&
	\multirow{2}{*}{$m$}&
	\multirow{2}{*}{$L$}&
	\multicolumn{3}{c}{B-P}&
	\multicolumn{3}{c}{B-C}&
	\multicolumn{3}{c}{B-P-2R}&
	\multicolumn{3}{c}{B-C-P}&
	\multicolumn{3}{c}{Our algorithm}\\
	\cmidrule(l){5-7}\cmidrule(l){8-10}\cmidrule(l){11-13}\cmidrule(l){14-16}\cmidrule(l){17-19}&
	& & &$UB$&
	$LB$&
	$CPU$&
	$UB$&
	$LB$&
	$CPU$&
	$UB$&
	$LB$&
	$CPU$&
	$UB$&
	$LB$&
	$CPU$&
	$UB$&
	$LB$&
	$CPU$\\
	\midrule
	
	\endfirsthead
	
	\multicolumn{19}{c}{{\tablename} \thetable{} -- continued from previous page} \\
	
	\toprule
	\multirow{2}{*}{$Instance$}&
	\multirow{2}{*}{$n$}&
	\multirow{2}{*}{$m$}&
	\multirow{2}{*}{$L$}&
	\multicolumn{3}{c}{B-P}&
	\multicolumn{3}{c}{B-C}&
	\multicolumn{3}{c}{B-P-2R}&
	\multicolumn{3}{c}{B-C-P}&
	\multicolumn{3}{c}{Our algorithm}\\
	\cmidrule(l){5-7}\cmidrule(l){8-10}\cmidrule(l){11-13}\cmidrule(l){14-16}\cmidrule(l){17-19}&
	& & &$UB$&
	$LB$&
	$CPU$&
	$UB$&
	$LB$&
	$CPU$&
	$UB$&
	$LB$&
	$CPU$&
	$UB$&
	$LB$&
	$CPU$&
	$UB$&
	$LB$&
	$CPU$\\
	\midrule
	
	\endhead
	
	\bottomrule
	\multicolumn{19}{c}{continued on next page}
	\endfoot
	
	\bottomrule
	\endlastfoot
	$p1.2.p$&$30$&$2$&$37.5$&$250$&$2926$&$\sim$&$250$&$250$&$27$&$250$&$250$&$15$&$250$&$250$&$16$&$250$&$250$&$7$\\
	$p1.2.q$&$30$&$2$&$40$&$-$&$-$&$\sim$&$265$&$265$&$139$&$265$&$265$&$78$&$265$&$265$&$80$&$265$&$265$&$5$\\
	$p1.2.r$&$30$&$2$&$42.5$&$-$&$-$&$\sim$&$280$&$280$&$33$&$280$&$280$&$555$&$280$&$280$&$566$&$280$&$280$&$4$\\
	$p3.2.l$&$31$&$2$&$35$&$605$&$-$&$4737$&$590$&$590$&$53$&$605$&$590$&$59$&$591$&$-$&$2783$&$590$&$590$&$28$\\
	$p3.2.m$&$31$&$2$&$37.5$&$-$&$-$&$\sim$&$620$&$620$&$58$&$630.769$&$620$&$192$&$623.953$&$-$&$7121$&$620$&$620$&$33$\\
	$p3.2.n$&$31$&$2$&$40$&$-$&$-$&$\sim$&$660$&$660$&$48$&$662.453$&$660$&$1751$&$660$&$660$&$4345$&$660$&$660$&$28$\\
	$p3.2.o$&$31$&$2$&$42.5$&$-$&$-$&$\sim$&$690$&$690$&$46$&$699.444$&$690$&$811$&$699.444$&$-$&$73$&$690$&$690$&$19$\\
	$p3.2.p$&$31$&$2$&$45$&$-$&$-$&$\sim$&$720$&$720$&$74$&$730$&$720$&$3881$&$730$&$-$&$282$&$720$&$720$&$24$\\
	$p3.2.q$&$31$&$2$&$47.5$&$-$&$-$&$\sim$&$760$&$760$&$20$&$763.2$&$760$&$1497$&$763.2$&$-$&$1779$&$760$&$760$&$12$\\
	$p3.2.r$&$31$&$2$&$50$&$-$&$-$&$\sim$&$790$&$790$&$15$&$790$&$790$&$1253$&$790$&$790$&$1660$&$790$&$790$&$8$\\
	$p3.2.s$&$31$&$2$&$52.5$&$-$&$-$&$\sim$&$800$&$800$&$7$&$800$&$800$&$60$&$800$&$800$&$234$&$800$&$800$&$0$\\
	$p3.3.s$&$31$&$3$&$35$&$738.913$&$416$&$\sim$&$720$&$720$&$384$&$738.913$&$720$&$5136$&$729.36$&$-$&$5004$&$720$&$720$&$90$\\
	$p3.3.t$&$31$&$3$&$36.7$&$763.688$&$4181$&$\sim$&$760$&$760$&$257$&$763.688$&$760$&$157$&$760.693$&$-$&$2933$&$760$&$760$&$42$\\
	$\textbf{p4.2.f}$&$98$&$2$&$50$&$-$&$-$&$\sim$&$-$&$-$&$\sim$&$-$&$-$&$\sim$&$-$&$-$&$\sim$&$687$&$687$&$6550$\\
	$p4.2.h$&$98$&$2$&$60$&$-$&$-$&$\sim$&$835$&$835$&$2784$&$-$&$-$&$\sim$&$-$&$-$&$\sim$&$835$&$835$&$3125$\\
	$p4.2.i$&$98$&$2$&$65$&$-$&$-$&$\sim$&$918$&$918$&$5551$&$-$&$-$&$\sim$&$-$&$-$&$\sim$&$918$&$918$&$1064$\\
	$\textbf{p4.2.j}$&$98$&$2$&$70$&$-$&$-$&$\sim$&$969$&$965$&$\sim$&$-$&$-$&$\sim$&$-$&$-$&$\sim$&$965$&$965$&$2777$\\
	$\textbf{p4.2.k}$&$98$&$2$&$75$&$-$&$-$&$\sim$&$1027$&$1022$&$\sim$&$-$&$-$&$\sim$&$-$&$-$&$\sim$&$1022$&$1022$&$2751$\\
	$\textbf{p4.2.l}$&$98$&$2$&$80$&$-$&$-$&$\sim$&$1080$&$1074$&$\sim$&$-$&$-$&$\sim$&$-$&$-$&$\sim$&$1074$&$1074$&$7172$\\
	$\textbf{p4.2.m}$&$98$&$2$&$85$&$-$&$-$&$\sim$&$1137$&$1132$&$\sim$&$-$&$-$&$\sim$&$-$&$-$&$\sim$&$1132$&$1132$&$4610$\\
	$\textbf{p4.2.r}$&$98$&$2$&$110$&$-$&$-$&$\sim$&$1293$&$1292$&$\sim$&$-$&$-$&$\sim$&$-$&$-$&$\sim$&$1292$&$1292$&$5016$\\
	$p4.2.t$&$98$&$2$&$120$&$-$&$-$&$\sim$&$1306$&$1306$&$5978$&$-$&$-$&$\sim$&$-$&$-$&$\sim$&$1306$&$1306$&$0$\\
	$p4.3.g$&$81$&$3$&$36.7$&$653$&$653$&$52$&$665$&$653$&$\sim$&$656.375$&$653$&$110$&$653$&$653$&$306$&$653$&$653$&$6587$\\
	$p4.3.h$&$90$&$3$&$40$&$729$&$729$&$801$&$761$&$729$&$\sim$&$735.375$&$599$&$\sim$&$730.704$&$-$&$3858$&$736$&$729$&$\sim$\\
	$p4.3.i$&$94$&$3$&$43.3$&$809$&$809$&$4920$&$830$&$809$&$\sim$&$813.625$&$766$&$\sim$&$809$&$809$&$2989$&$815$&$809$&$\sim$\\
	$p4.4.i$&$68$&$4$&$32.5$&$657$&$657$&$23$&$660$&$657$&$\sim$&$665.4$&$657$&$74$&$657$&$657$&$83$&$657$&$657$&$935$\\
	$p4.4.j$&$76$&$4$&$35$&$732$&$732$&$141$&$784$&$732$&$\sim$&$741.472$&$732$&$5138$&$732$&$732$&$589$&$755$&$732$&$\sim$\\
	$p4.4.k$&$83$&$4$&$37.5$&$821$&$821$&$558$&$860$&$821$&$\sim$&$831.945$&$816$&$\sim$&$821.803$&$-$&$4007$&$858$&$821$&$\sim$\\
	$p5.2.l$&$64$&$2$&$30$&$-$&$-$&$\sim$&$800$&$800$&$399$&$800$&$800$&$3$&$800$&$800$&$4$&$800$&$800$&$71$\\
	$p5.2.m$&$64$&$2$&$32.5$&$-$&$-$&$\sim$&$860$&$860$&$3865$&$860$&$860$&$32$&$860$&$860$&$38$&$860$&$860$&$90$\\
	$p5.2.n$&$64$&$2$&$35$&$-$&$-$&$\sim$&$930$&$925$&$\sim$&$930$&$925$&$89$&$925$&$925$&$1393$&$925$&$925$&$2373$\\
	$p5.2.o$&$64$&$2$&$37.5$&$-$&$-$&$\sim$&$1030$&$1020$&$\sim$&$1030$&$1020$&$271$&$1020$&$1020$&$2233$&$1025$&$1020$&$\sim$\\
	$p5.2.p$&$64$&$2$&$40$&$-$&$-$&$\sim$&$1150$&$1150$&$3955$&$1150$&$1150$&$657$&$1150$&$1150$&$727$&$1150$&$1150$&$77$\\
	$\textbf{p5.2.q}$&$64$&$2$&$42.5$&$-$&$-$&$\sim$&$1680$&$1195$&$\sim$&$-$&$-$&$\sim$&$-$&$-$&$\sim$&$1195$&$1195$&$6597$\\
	$p5.2.r$&$64$&$2$&$45$&$-$&$-$&$\sim$&$1680$&$1260$&$\sim$&$1260$&$1260$&$123$&$1260$&$1260$&$133$&$1300$&$1269$&$\sim$\\
	$p5.2.s$&$64$&$2$&$47.5$&$-$&$-$&$\sim$&$1365$&$1340$&$\sim$&$1340$&$1340$&$1072$&$1340$&$1340$&$845$&$1340$&$1340$&$3048$\\
	$p5.2.t$&$64$&$2$&$50$&$-$&$-$&$\sim$&$1400$&$1400$&$5136$&$1400$&$-$&$1297$&$1400$&$1400$&$4559$&$1400$&$1400$&$418$\\
	$p5.2.u$&$64$&$2$&$52.5$&$-$&$-$&$\sim$&$1510$&$1460$&$\sim$&$1460$&$1460$&$3488$&$1460$&$1460$&$4561$&$1460$&$1460$&$3263$\\
	$\textbf{p5.2.v}$&$64$&$2$&$55$&$-$&$-$&$\sim$&$1530$&$1520$&$\sim$&$1510$&$-$&$4462$&$1510$&$-$&$4948.16$&$1505$&$1505$&$3497$\\
	$\textbf{p5.2.w}$&$64$&$2$&$57.5$&$-$&$-$&$\sim$&$1680$&$1565$&$\sim$&$-$&$-$&$\sim$&$-$&$-$&$\sim$&$1565$&$1565$&$5875$\\
	$p5.2.x$&$64$&$2$&$60$&$-$&$-$&$\sim$&$1610$&$1610$&$1048$&$-$&$-$&$\sim$&$-$&$-$&$\sim$&$1610$&$1610$&$128$\\
	$\textbf{p5.2.y}$&$64$&$2$&$62.5$&$-$&$-$&$\sim$&$1655$&$1645$&$\sim$&$-$&$-$&$\sim$&$-$&$-$&$\sim$&$1645$&$1645$&$457$\\
	$p5.2.z$&$64$&$2$&$65$&$-$&$-$&$\sim$&$1680$&$1680$&$1604$&$-$&$-$&$\sim$&$-$&$-$&$\sim$&$1680$&$1680$&$0$\\
	$p5.3.l$&$64$&$3$&$20$&$595$&$595$&$33$&$615$&$595$&$\sim$&$605$&$595$&$31$&$600$&$595$&$35$&$615$&$595$&$\sim$\\
	$p5.3.m$&$64$&$3$&$21.7$&$650$&$650$&$2$&$660$&$650$&$\sim$&$650$&$650$&$1$&$650$&$650$&$1$&$660$&$650$&$\sim$\\
	$p5.3.n$&$64$&$3$&$23.3$&$755$&$755$&$42$&$765$&$755$&$\sim$&$755$&$755$&$3$&$755$&$755$&$3$&$765$&$755$&$\sim$\\
	$p5.3.q$&$64$&$3$&$28.3$&$-$&$-$&$\sim$&$1260$&$1070$&$\sim$&$1090$&$1070$&$521$&$1076.25$&$-$&$4694$&$1110$&$1070$&$\sim$\\
	$p5.3.t$&$64$&$3$&$33.3$&$-$&$-$&$\sim$&$1320$&$1260$&$\sim$&$1270$&$1260$&$5152$&$1270$&$-$&$16$&$1320$&$1260$&$\sim$\\
	$p5.3.u$&$64$&$3$&$35$&$-$&$-$&$\sim$&$1395$&$1345$&$\sim$&$1350$&$-$&$123$&$1350$&$-$&$149$&$1395$&$1345$&$\sim$\\
	$p5.4.l$&$44$&$4$&$15$&$430$&$430$&$1$&$445$&$430$&$\sim$&$430$&$430$&$0$&$430$&$430$&$0$&$430$&$430$&$2077$\\
	$p5.4.m$&$52$&$4$&$16.2$&$555$&$555$&$0$&$560$&$555$&$\sim$&$555$&$555$&$0$&$555$&$555$&$0$&$555$&$555$&$1357$\\
	$p5.4.n$&$60$&$4$&$17.5$&$620$&$620$&$0$&$640$&$620$&$\sim$&$620$&$620$&$0$&$620$&$620$&$0$&$620$&$620$&$7048$\\
	$p5.4.o$&$60$&$4$&$18.8$&$690$&$690$&$1$&$720$&$690$&$\sim$&$690$&$690$&$0$&$690$&$690$&$0$&$720$&$690$&$\sim$\\
	$p5.4.p$&$64$&$4$&$20$&$765$&$765$&$729$&$820$&$765$&$\sim$&$790$&$765$&$1238$&$775.714$&$765$&$1372$&$820$&$765$&$\sim$\\
	$p5.4.q$&$64$&$4$&$21.2$&$860$&$860$&$1$&$880$&$860$&$\sim$&$860$&$860$&$2$&$860$&$860$&$2$&$880$&$860$&$\sim$\\
	$p5.4.v$&$64$&$4$&$27.5$&$1320$&$1320$&$446$&$1340$&$1320$&$\sim$&$1320$&$1320$&$12$&$1320$&$1320$&$12$&$1340$&$1320$&$\sim$\\
	$p5.4.y$&$64$&$4$&$31.2$&$-$&$-$&$\sim$&$1620$&$1520$&$\sim$&$1520$&$1455$&$\sim$&$1520$&$1520$&$46$&$1620$&$1520$&$\sim$\\
	$p5.4.z$&$64$&$4$&$32.5$&$-$&$-$&$\sim$&$1680$&$1620$&$\sim$&$1620$&$1620$&$550$&$1620$&$1620$&$562$&$1680$&$1620$&$\sim$\\
	$p6.2.j$&$62$&$2$&$30$&$-$&$-$&$\sim$&$948$&$948$&$2393$&$948$&$948$&$139$&$948$&$948$&$149$&$948$&$948$&$1338$\\
	$p6.2.k$&$62$&$2$&$32.5$&$-$&$-$&$\sim$&$1032$&$1032$&$4016$&$1032$&$1032$&$223$&$1032$&$1032$&$244$&$1032$&$1032$&$699$\\
	$p6.2.l$&$62$&$2$&$35$&$-$&$-$&$\sim$&$1116$&$1116$&$3828$&$1116$&$1116$&$5699$&$1116$&$1116$&$6471$&$1116$&$1116$&$39$\\
	$p6.2.m$&$62$&$2$&$37.5$&$-$&$-$&$\sim$&$1188$&$1188$&$1442$&$-$&$-$&$\sim$&$-$&$-$&$\sim$&$1188$&$1188$&$680$\\
	$p6.2.n$&$62$&$2$&$40$&$-$&$-$&$\sim$&$1260$&$1260$&$1473$&$-$&$-$&$\sim$&$-$&$-$&$\sim$&$1260$&$1260$&$1$\\
	$p6.3.m$&$62$&$3$&$25$&$1104$&$-$&$33$&$1080$&$1080$&$1175$&$1104$&$-$&$20$&$1094.1$&$-$&$6407$&$1080$&$1080$&$432$\\
	$p7.2.g$&$87$&$2$&$70$&$-$&$-$&$\sim$&$459$&$459$&$1226$&$459$&$459$&$44$&$459$&$459$&$58$&$459$&$459$&$589$\\
	$p7.2.h$&$92$&$2$&$80$&$-$&$-$&$\sim$&$523$&$521$&$\sim$&$521$&$521$&$5101$&$521$&$521$&$6327$&$521$&$521$&$1977$\\
	$\textbf{p7.2.i}$&$98$&$2$&$90$&$-$&$-$&$\sim$&$585$&$580$&$\sim$&$-$&$-$&$\sim$&$-$&$-$&$\sim$&$580$&$580$&$6271$\\
	$\textbf{p7.2.t}$&$100$&$2$&$200$&$-$&$-$&$\sim$&$1181$&$1179$&$\sim$&$-$&$-$&$\sim$&$-$&$-$&$\sim$&$1179$&$1179$&$6934$\\
	$p7.3.h$&$59$&$3$&$53.3$&$425$&$425$&$8$&$436$&$425$&$\sim$&$429$&$425$&$3$&$425$&$425$&$13$&$425$&$425$&$4461$\\
	$p7.3.i$&$70$&$3$&$60$&$487$&$487$&$3407$&$535$&$487$&$\sim$&$496.976$&$487$&$436$&$488.5$&$487$&$3357$&$509$&$487$&$\sim$\\
	$p7.3.j$&$80$&$3$&$66.7$&$570.5$&$2654$&$\sim$&$611$&$564$&$\sim$&$570.5$&$564$&$4207$&$564$&$564$&$4289$&$573$&$564$&$\sim$\\
	$p7.3.k$&$91$&$3$&$73.3$&$-$&$-$&$\sim$&$688$&$633$&$\sim$&$633.182$&$633$&$1173$&$633$&$633$&$2751$&$655$&$633$&$\sim$\\
	$p7.3.m$&$96$&$3$&$86.7$&$-$&$-$&$\sim$&$1374$&$762$&$\sim$&$762$&$762$&$928$&$762$&$762$&$1202$&$817$&$762$&$\sim$\\
	$p7.3.n$&$99$&$3$&$93.3$&$-$&$-$&$\sim$&$900$&$820$&$\sim$&$820$&$820$&$2300$&$820$&$820$&$3034$&$889$&$820$&$\sim$\\
	$p7.4.j$&$51$&$4$&$50$&$462$&$462$&$1$&$481$&$462$&$\sim$&$462$&$462$&$2$&$462$&$462$&$2$&$465$&$462$&$\sim$\\
	$p7.4.k$&$61$&$4$&$55$&$520$&$520$&$73$&$586$&$520$&$\sim$&$524.607$&$520$&$96$&$520$&$520$&$91$&$541$&$520$&$\sim$\\
	$p7.4.l$&$70$&$4$&$60$&$590$&$590$&$778$&$667$&$590$&$\sim$&$593.625$&$590$&$576$&$590$&$590$&$173$&$632$&$590$&$\sim$\\
	$p7.4.n$&$87$&$4$&$70$&$-$&$-$&$\sim$&$809$&$730$&$\sim$&$730$&$730$&$85$&$730$&$730$&$95$&$803$&$730$&$\sim$\\
	$p7.4.o$&$91$&$4$&$75$&$-$&$-$&$\sim$&$909$&$781$&$\sim$&$786.762$&$781$&$4434$&$784.676$&$-$&$6492$&$903$&$781$&$\sim$\\
\label{tab:comparison_literature}
\end{longtable}
\end{landscape}}
\end{small}

Next, we compare the performance of all the exact methods on a per-data-set basis. Table \ref{tab:summary} summarizes the numbers of instances being solved by each method. We did not report the $CPU$ time in this table because of some missing informations from the other methods.

\begin{table}[h]
	\caption{Comparison between the numbers of instances being solved by the exact methods in the literature.}\label{tab:summary} 
	\begin{center}
		\begin{tabular}{cccccc}
			\toprule
			Set&B-P&B-C&B-P-2R&B-C-P&Our algorithm\\
			\midrule
			$1$&$51/54$&$\textbf{54/54}$&$\textbf{54/54}$&$\textbf{54/54}$&$\textbf{54/54}$\\
			$2$&$\textbf{33/33}$&$\textbf{33/33}$&$\textbf{33/33}$&$\textbf{33/33}$&$\textbf{33/33}$\\
			$3$&$50/60$&$\textbf{60/60}$&$\textbf{60/60}$&$51/60$&$\textbf{60/60}$\\
			$4$&$25/60$&$22/60$&$20/60$&$22/60$&$\textbf{30/60}$\\
			$5$&$48/78$&$44/78$&$\textbf{60/78}$&$59/78$&$54/78$\\
			$6$&$36/42$&$\textbf{42/42}$&$36/42$&$38/42$&$\textbf{42/42}$\\
			$7$&$27/60$&$23/60$&$\textbf{38/60}$&$34/60$&$27/60$\\
			\midrule
			$Total$&$270/387$&$278/387$&$301/387$&$291/387$&$300/387$\\
			\bottomrule
		\end{tabular}
	\end{center}
\end{table}

A first remark from these results is that instances with large values of $L$ and $m$ are generally more difficult to solve than those with smaller values. This can be clearly observed with our method on the data sets $4$, $5$ and $7$. On the other hand, none of the exact methods had a difficulty to solve the instances of the sets $1$, $2$, and $3$, because these instances have a small numbers of accessible customers. The random distribution of customers around the depots could also make the optimal solutions easier to locate. Only a minor exception was noticed for B-P and B-C-P on some instances of the set $3$.

The random distribution of customers is also the case for the sets $4$ and $7$, however their numbers of accessible customers are larger than those of the first three sets, e.g. they can reach $100$. These large numbers cause a difficulty for all the exact methods to solve the corresponding instances. Particularly, the number of solved instances did not exceed $58$ out of the $120$ instances by any of the five methods.

Finally, the instances of the sets $5$ and $6$ contain a special geometric structure. These instances have no more than $64$ accessible customers, which are arranged on a grid in a way that those with large profits are far away from the depots. It appears that these instances are difficult to solve. This is especially the case with B-P and B-C algorithms. The B-P-2R and the B-C-P algorithms of \citet{bib:Keshtkarana15} only had problems with the set $6$, while they obtained the best results for the set $5$. However, our cutting-plane algorithm obtained quite good results for these two sets. It was able to solve all the instances of the set $6$ and most of the instances in the set $5$. We had only few difficulties with the set $5$, and more precisely on some instances with $4$ available vehicles. A closer look into the execution of the algorithm on those few instances revealed to us that the CPA only made progress in improving the incumbent solution or finding equivalent ones, while it was not much reducing the upper bound.

To summarize, our algorithm was able to prove the optimality of all the instances of the sets $1$, $2$, $3$, and $6$ and a large number of instances from the other three sets. Although the instances of the set $4$ are the hardest ones to solve, our CPA was able to prove the optimality of $30$ out of the $60$ instances, while none of the existing algorithms was able to reach that number for this set. In total, the proposed approach is capable of solving \nbsolved{} out of the $387$ instances.

\begin{table}[h]
	\caption{Comparison between each two exact methods apart.}\label{tab:comparison_literature_two} 
	\begin{center}
		\begin{tabular}{lccccc}
			\toprule
			&B-P&B-C&B-P-2R&B-C-P&CPA\\
			\midrule
			B-P &$-$&$21$&$8$&$6$&$15$\\
			B-C &$29$&$-$&$13$&$19$&$0$\\
			B-P-2R &$39$&$36$&$-$&$16$&$26$\\
			B-C-P &$27$&$32$&$6$&$-$&$22$\\
			CPA&$45$&$22$&$25$&$31$&$-$\\
			\bottomrule
		\end{tabular}
	\end{center}
\end{table}

For further comparison between the exact methods in the literature, we present in table \ref{tab:comparison_literature_two} a comparison between each two methods apart, by giving the number of instances being solved by one of the two methods and not by the other one. Each cell of this table reports the number of instances being solved by the method present in its row but not by the method in the column. From the results shown in this table, we can see that the number of instances being distinctively solved by our method is $45$ compared to the B-P algorithm, $22$ compared to the B-C algorithm, and respectively $25$ and $31$ compared to the B-P-2R and the B-C-P algorithms.

Moreover, we can notice from Table \ref{tab:comparison_literature} that our CPA was able to improve the upper bounds of respectively $32$ and $27$ instances more than the two algorithms of \citet{bib:Keshtkarana15}. Overall, our approach is clearly efficient and competitive with the existing methods in the literature. We were able to prove the optimality of \nbnewclosed{} instances that have been unsolved in the literature. These instances are marked in bold in Table \ref{tab:comparison_literature}.

\section*{Conclusion and future work}\label{sec:conclusion}
The Team Orienteering Problem is one of the well known variants of the Vehicle Routing Problem with Profits. In this article, we presented a new exact algorithm to solve this problem based on a cutting-plane approach. Several types of cuts are proposed in order to strengthen the classical linear formulation. The corresponding cuts are generated and added to the model during the solving process. They include symmetry breaking, generalized subtour eliminations, boundaries on profits/numbers of customers, forcing mandatory customers, removing irrelevant components and clique and independent-set cuts based on graph of incompatibilities between variables. The experiments conducted on the standard benchmark of TOP confirm the effectiveness of our approach. Our algorithm is able to solve a large number and a large variety of instances, some of those instances have been unsolved in the literature.

Interestingly, the branch-and-price algorithm of \citet{bib:Boussier07} and our Cutting Plane algorithm has complementary performance to each other.
This gives us a hint that further development of a Branch-and-Cut-and-Price type of algorithm which incorporates our presented ideas is a promising direction towards improving the solving method for TOP. We also plan to adapt the presented approach to meet new challenges. Those could include variants of TOP on arcs, such as the Team Orienteering Arc Routing Problem (TOARP) which was addressed in \citep{bib:Archetti13}. On the other hand, by taking into consideration the time scheduling of the visits, the CPA can be extended to solve other variants of TOP and VRP, such as the Team Orienteering Problem with Time Windows and/or Synchronization Constraints \citep[e.g.,][]{bib:Labadi12,bib:Souffriau13,bib:Guibadj13,bib:Afifi16}.

\section*{Acknowledgement}
This work is carried out in the framework of the Labex MS2T, which was funded by the French Government, through the program “Investments for the future” managed by the National Agency for Research (Reference ANR-11-IDEX-0004-02). It is also partially supported by the Regional Council of Picardie under TOURNEES SELECTIVES project and TCDU project (Collaborative Transportation in Urban Distribution, ANR-14-CE22-0017).


\bibliographystyle{abbrvnat}
\bibliography{bibliography}

\end{document}